# Enabling Embedded Inference Engine with the ARM Compute Library: A Case Study

Dawei Sun, Shaoshan Liu*, and Jean-Luc Gaudiot

If you need to enable deep learning on low-cost embedded SoCs, should you port an existing deep learning framework or should you build one from scratch? In this paper, we seek to answer this question by sharing our practical experience of building an embedded inference engine using the ARM Compute Library (ACL). The results show that, contradictory to conventional wisdom, for simple models, it takes much less development time to build an inference engine from scratch as opposed to porting existing frameworks. In addition, by utilizing ACL, we managed to build an inference engine that outperforms TensorFlow by 25%. Our conclusion is that, with embedded devices, we most likely will use very simple deep learning models for inference, and with well-developed building blocks such as ACL, it may yield better performance and result in lower development time if the engine is built from scratch.

## Enabling Inference on Embedded Devices

We were building an internet-of-things product with inference capabilities on our bare-metal ARM SoC, code-named Zuluko (the Zuluko SoC contains four ARM v7 cores running at 1 GHz, as well as 512 MB of RAM). At its peak it consumes about 3 W of power and costs only about four dollars. Everything was progressing smoothly until we had to enable high-performance inference capabilities on it. An easy option was to migrate an existing deep learning platform, so we chose to migrate TensorFlow [1] since it delivered the best performance on ARM-Linux platforms based on our study.

We thought this would be an easy task, but it took us days to port all the dependencies of TensorFlow before we could even run the TensorFlow platform. Eventually, after a week of intensive efforts, we managed to run TensorFlow on Zuluko. This experience made us wonder whether it could be worthwhile to build a platform from scratch or better to port an existing platform. This question had two implications: first, without basic building blocks such as convolution operator, it would be very hard to build an inference engine from scratch. Second, an inference engine built-from-scratch may not outperform a well-tested deep learning framework. Let us examine these problems in the coming sections.

## Building Inference Engine with the ARM Compute Library

Recently, ARM announced their Compute Library [2], a comprehensive collection of software functions implemented for the ARM Cortex-A family of CPU processors and the ARM Mali family of GPUs. Specifically, it provides the basic building blocks for Convolutional Neural Networks, including Activation, Convolution, Fully Connected, Locally Connected, Normalization, Pooling, and Soft-Max. These are exactly what we needed to build an inference engine. We went ahead and attempted to build a SqueezeNet [3] using these building blocks.

To construct SqueezeNet, we started by building the fire module proposed in [3]. As shown in Figure 1, SqueezeNet utilizes a 1 X 1 convolution kernel to reduce the input size of the 3 X 3 convolution layer while maintaining similar inference accuracy. Then SqueezeNet utilizes an expand strategy to guarantee the dimension of the network does not change. This is the fire module and it is the core of SqueezeNet. We utilized the ACL core operators to implement the fire module and our implementation eliminates the need for extra memory copy otherwise needed for concatenation operation.

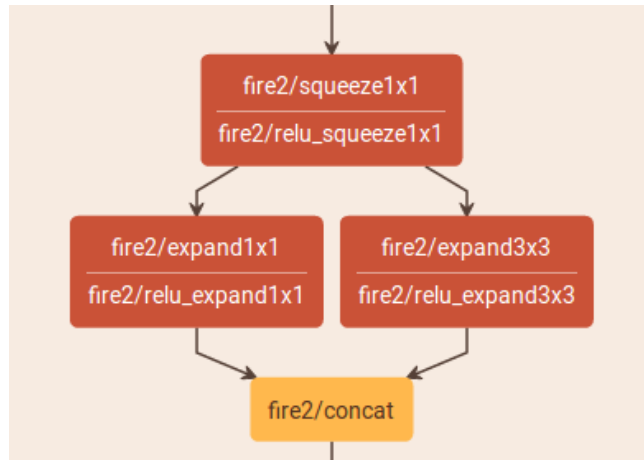

Figure 1: The fire layer

After going through several fire modules, we reach the output layer of SqueezeNet. As shown in Figure 2, it includes a dropout operation and a global pooling operation. Since ACL currently does not support dropout and global pooling, we implemented our own operators from scratch. For a dropout operator, the main purpose is to prevent overfitting in training by randomly setting some outputs to zero, since we use the network for inference we can basically eliminate this layer. But to compensate for the change in output, we added an attenuation coefficient after pool10 layer to match the attenuation introduced in the original dropout layer.

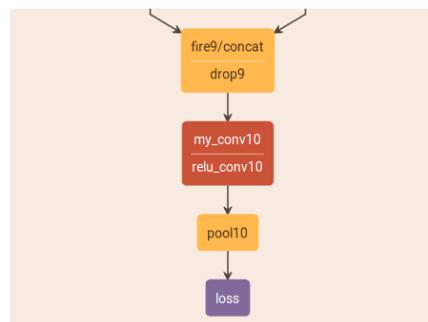

Figure 2: SqueezeNet implementation

## Performance

Now that we have built a SqueezeNet engine using the basic building blocks from ACL, we can delve into the performance of TensorFlow versus that of ACL. To ensure a fair comparison, we enabled ARM NEON vector computation optimization on TensorFlow and we also chose to use NEON-enabled building blocks when building our SqueezeNet engine. By making sure both engines utilize the NEON vector computation, we ensured that any performance difference would have been caused exclusively by the platform itself.

As shown in Figure 3, we ran SqueezeNet with TensorFlow on our Zuluko platform, using four cores. On average; it took 420 ms to process a 227 X 227 RGB image. Using the SqueezeNet engine built from ACL, it took only 320 ms to process the same image, hence a 25% speedup. To better understand the source of the gain in performance, we moved one step further and divided the processing time into two groups: the first group included convolution, RELU, and concatenate, while the second group included pooling and soft-max. The breakdown results show that our SqueezeNet engine outperforms TensorFlow by 23% in group 1 and 110% in group 2. As for resource utilization, when running on TensorFlow, the average CPU usage is 75% while the average memory usage is about 9 MB. At the same time, when running on ACL, the average CPU usage is 90% and the average memory usage is about 10 MB.

There could be two reasons for the performance improvement: first, ACL provides better NEON optimization since all the operators in ACL were developed using NEON intrinsic operators directly whereas TensorFlow relied on the

ARM compiler to provide NEON optimization. Second, TensorFlow platform itself likely introduces some performance overhead.

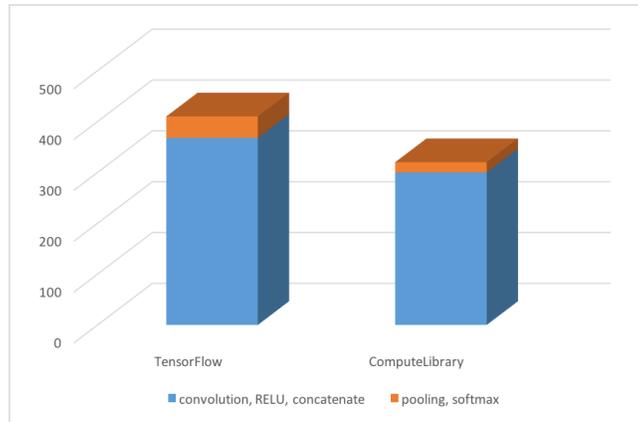

**Figure 3: TensorFlow *vs.* ACL**

Next we sought to squeeze additional performance from TensorFlow and undertook to check whether it could outperform the inference engine built on top of ACL. We attempted this by using the vector quantization optimization technique [4]. The gist of this optimization is to use 8-bit weights so as to trade accuracy for performance. In addition, with 8-bit weights, we can use vector computations to process multiple data units with one instruction. However, this optimization comes at a cost: it introduces re-quantize and de-quantize operations. In Figure 4, we compare the performance of two cases: quantization *vs.* without quantization. By using vector quantization, we managed to improve the performance of convolution by 25%. However, it introduces significant overhead because of the de-quantization and re-quantization operations. Overall, it actually slows down the whole inference process by more than 100 ms.

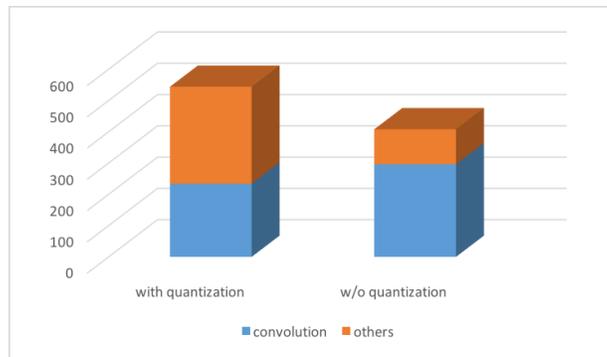

**Figure 4: quantization performance**

## Discussions

In this paper we shared our experience with building a deep learning inference engine from scratch by using ACL. What led to the decision of building such an engine was the difficulty in migrating an existing engine and its disappointing performance. Existing deep learning engines are built for generality, suitable for both training and inference tasks. Often, these engines are not optimized for embedded inference tasks. Also, these engines depend on many other third party libraries not readily available on "bare-metal" embedded systems, making them very hard to migrate. In contrast, speaking of performance, with building blocks provided by ACL, we could build embedded inference engines to deliver high performance since we could fully utilize the heterogeneous computing resources provided by the SoC. Thus, the question becomes whether it is easier to migrate existing engines, or is it easier to build from scratch? Our experience presented in this paper shows that if the model is simple, it is much easier to build from scratch. As the model gets increasingly complicated, we may hit a point where it is more efficient to port an existing engine. However, it is highly unlikely that we would need to use highly complicated models for embedded inference tasks. Therefore, we conclude that built-from-scratch embedded inference engine may be a viable solution to bring deep learning capabilities to embedded devices.

**Dawei Sun** is currently with Tsinghua University and PerceptIn, working on Deep Learning and cloud infrastructures, autonomous robots, as well as embedded systems. Contact him at: sdw14@mails.tsinghua.edu.cn

**Dr. Shaoshan Liu\*** is the corresponding author of this paper. He is currently the Co-Founder and Chairman of PerceptIn, working on developing the next-generation robotics platform. Contact him at: shaoshan.liu@perceptin.io

**Dr. Jean-Luc Gaudiot** is professor in the Electrical Engineering and Computer Science Department at the University of California, Irvine and is currently serving as the 2017 President of the IEEE Computer Society. Contact him at gaudiot@uci.edu